\documentclass[10pt,twocolumn,letterpaper]{article}

\usepackage{cvpr}
\usepackage{times}
\usepackage{epsfig}
\usepackage{graphicx}
\usepackage{amsmath}
\usepackage{amssymb}
\usepackage{booktabs}
\usepackage {multirow}

\usepackage[pagebackref=true,breaklinks=true,letterpaper=true,colorlinks,bookmarks=false]{hyperref}
\newcommand{\tabincell}[2]{\begin{tabular}{@{}#1@{}}#2\end{tabular}}

\cvprfinalcopy 

\def\cvprPaperID{3286} 

\ifcvprfinal\fi
\begin{document}

\title{PAD-Net: Multi-Tasks Guided Prediction-and-Distillation Network \\for Simultaneous Depth Estimation and Scene Parsing}
\author{Dan Xu$^{1}$, \quad Wanli Ouyang$^2$, \quad Xiaogang Wang$^3$, \quad Nicu Sebe$^1$\vspace{5pt} \\
$^1$The University of Trento, $^2$The University of Sydney, $^3$The Chinese University of Hong Kong\\
{\tt\small \{dan.xu, niculae.sebe\}@unitn.it \ wanli.ouyang@sydney.edu.au \ xgwang@ee.cuhk.edu.hk}
}

\maketitle

\begin{abstract}
Depth estimation and scene parsing are two particularly important tasks in visual scene understanding. In this paper we tackle the problem of simultaneous depth estimation and scene parsing in a joint CNN. The task can be typically treated as a deep multi-task learning problem~\cite{ranjan2016hyperface}. Different from previous methods directly optimizing multiple tasks given the input training data, this paper proposes a novel multi-task guided prediction-and-distillation network (PAD-Net), which first predicts a set of intermediate auxiliary tasks ranging from low level to high level, and then the predictions from these intermediate auxiliary tasks are utilized as multi-modal input via our proposed multi-modal distillation modules for the final tasks. During the joint learning, the intermediate tasks not only act as supervision for learning more robust deep representations but also provide rich multi-modal information for improving the final tasks. Extensive experiments are conducted on two challenging datasets (\ie~NYUD-v2 and Cityscapes) for both the depth estimation and scene parsing tasks, demonstrating the effectiveness of the proposed approach.
\end{abstract}

\section{Introduction}
\begin{figure}[!t]
\centering
\includegraphics[width=3in, height=3.2in]{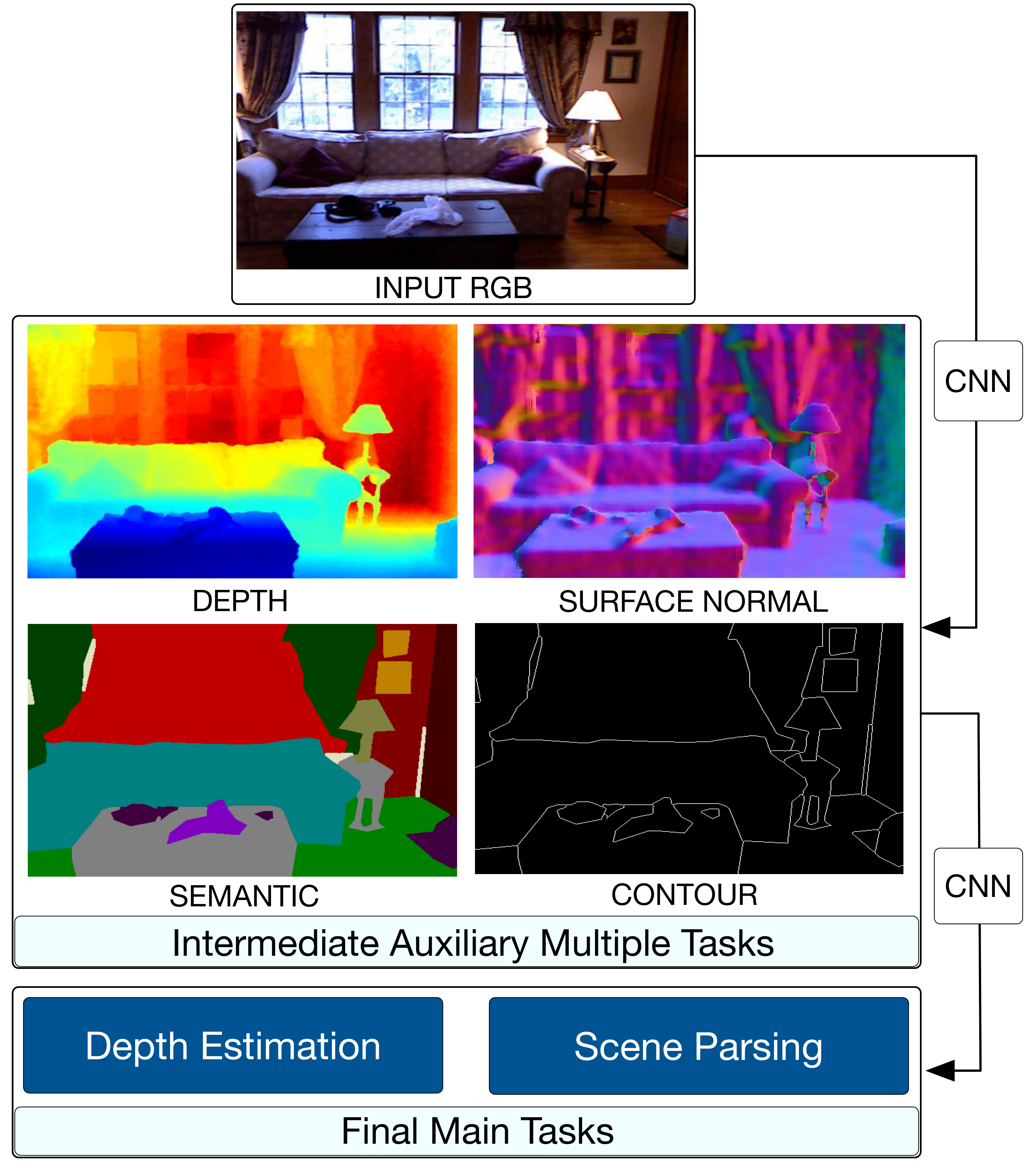} 
\caption{Motivation illustration. The proposed approach utilizes multiple intermediate multi-modal output from multi-task predictions as guidance to facilitate the final main-tasks. Different intermediate tasks ranging from low level to high level are considered, \ie~monocular depth prediction, surface normal estimation, contour prediction and semantic parsing. }
\label{motivation}
\vspace{-15pt}
\end{figure}

Depth estimation and scene parsing are both fundamental tasks for visual scene perception and understanding. Significant efforts have been made by many researchers on the two tasks in recent years. Due to the powerful deep learning technologies, the performance of the two individual tasks has been greatly improved~\cite{eigen2014depth, xu2017multi, chen2016deeplab}. 
Since these two tasks are correlated, jointly learning a single network for the two tasks is a promising research line. 


\par Typical deep multi-task learning approaches mainly focused on the final prediction level via employing the cross-modal interactions to mutually refining the tasks~\cite{jafari2017analyzing, wang2015towards} or designing more effective joint-optimization objective functions~\cite{mousavian2016joint, kendall2017multi}. These methods directly learn to predict the two tasks given the same input training data. Under this setting, they usually require the deep models to partially share network parameters or hidden representations. However, simultaneously learning the different tasks using distinct loss functions makes the network optimization complicated, and it is generally not easy to obtain a good generalization ability for all the tasks, which therefore brings worse performance on some of the tasks compared with the optimization with only a single task, as found by UberNet~\cite{kokkinos2017ubernet}.
In this paper, we explore multi-task deep learning from a different direction, \ie~using intermediate multi-task outputs as multi-modal input data. This is motivated by three observations. First, it is well-known that multi-modal data improve the performance of deep predictions. Take the task of scene parsing as an example, a CNN trained with RGB-D data should perform better than the CNN trained with only the RGB data. If we do not have the depth data available, we can use a CNN to predict the depth maps and then use them as input. Second, instead of using the output only from the target tasks, \ie~semantic and depth maps, as the multi-modal input, the powerful CNN is able to predict more information related, such as contour and surface normal. Third, how to effectively use the multi-modal data obtained from intermediate auxiliary predictions to facilitates the final tasks is particularly important. In other words, it is a crucial point that how to design a good network architecture so that the network communicates or shares information based on the multi-modal data for different tasks, while other deep multi-task learning models such as Cross-stitch Net~\cite{misra2016cross}, Sluice Net~\cite{ruder2017sluice}, and Deep Relation Net~\cite{long2015learning}, assume only single-modal data and thus do not consider it. 
\par Based on the observations above, a multi-tasks guided prediction-and-distillation network (PAD-Net) is proposed. Specifically, we first learn to use a front-end deep CNN and the input RGB data to produce a set of intermediate auxiliary tasks (see Fig.~\ref{motivation}). The auxiliary tasks range from low level to high level involving two continuous regression tasks (monocular depth prediction and surface normal estimation) and two discrete classification tasks (scene parsing and contour detection). The produced multiple predictions, \ie~depth maps,  surface normal, semantic maps and object contours, are then utilized as the multi-modal input of the next sub-deep-network for the final two main tasks. By involving an intermediate multi-task prediction module, the proposed PAD-Net not only adds deep supervision for optimizing the front-end network more effectively, but also is able to incorporate more knowledge from relevant domains. Since the predicted multi-modal results are highly complementary, we further propose multi-modal distillation strategies to better using these data. When the optimization of the whole PAD-Net is finished, the inference is only based on the RGB input.
\par To summarize, the contribution of this paper is threefold: 
(i) First, we propose a new multi-tasks guided prediction-and-distillation network (PAD-Net) structure for simultaneous depth estimation and scene parsing. It produces a set of intermediate auxiliary tasks providing rich multi-modal data for learning the target tasks. Although PAD-Net takes only RGB data as input, it is able to incorporate multi-modal information for improving the final tasks.
(ii) Second, we design and investigate three different multi-modal distillation modules for deep multi-modal data fusion, which we believe can be also applied in other scenarios such as multi-scale deep feature fusion. 
(iii) Third, extensive experiments on the challenging NYUD-v2 and Cityscapes datasets demonstrate the effectiveness of the proposed approach. Our approach achieves state-of-the-art results on NYUD-v2 on both the depth estimation and the scene parsing tasks, and obtains very competitive performance on the Cityscapes scene parsing task. More importantly, the proposed approach remarkably outperforms state-of-the-arts working on jointly optimizing both tasks. 

\begin{figure*}[!t]
\centering
\includegraphics[width=6.75in]{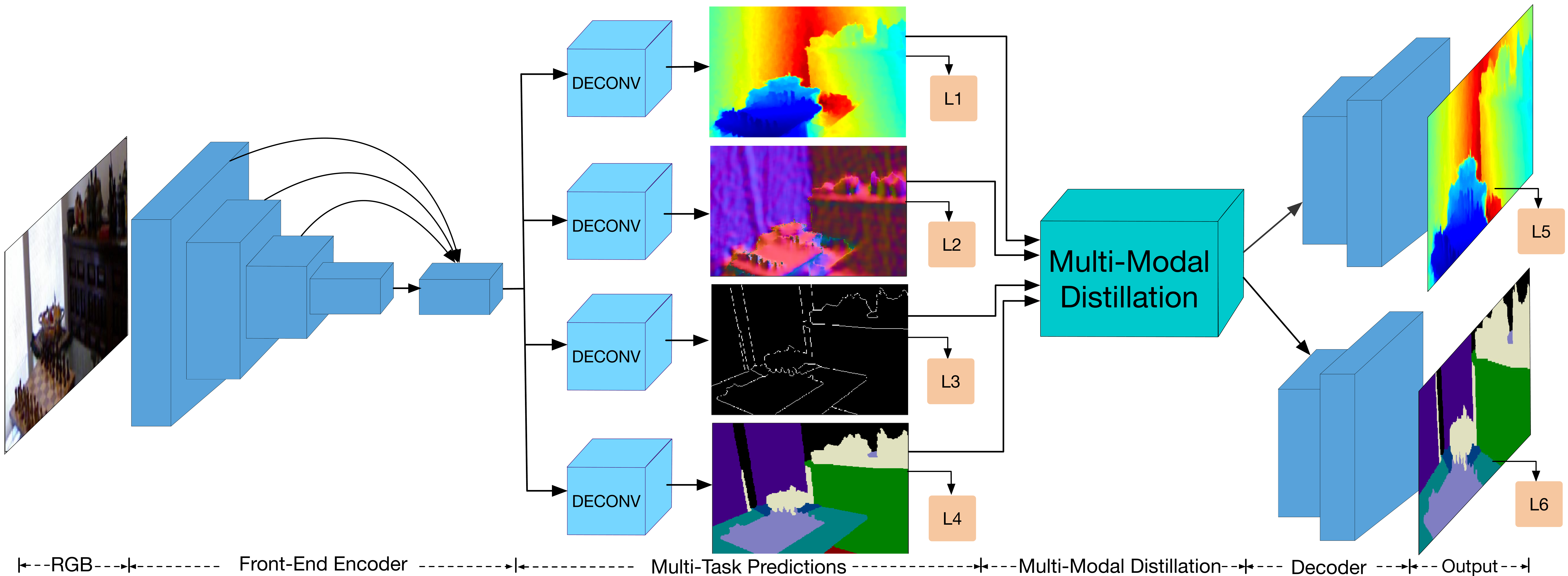} 
\caption{Illustration of the proposed PAD-Net for simultaneous depth estimation and scene parsing. The symbols of $\textrm{L1}$ to $\textrm{L6}$ denote different optimization losses for different tasks. `DECONV' denotes the deconvolutional operation for upsampling and generating task-specific feature maps. The cube `Multi-Modal Distillation' represents the proposed multi-modal distillation module for fusing the multiple predictions to improve the final main tasks.}
\label{framework}
\vspace{-15pt}
\end{figure*}
\vspace{-5pt}
\section{Related Work}
\vspace{-5pt}
\par\textbf{Depth estimation and scene parsing.}
The works on monocular depth estimation can be mainly grouped into two categories. The first group comprises the methods based on the hand-crafted features and graphical models~\cite{delage2006dynamic, saxena20083, liu2014discrete}. For instance, Saxena~\etal~\cite{saxena20083} proposed a discriminatively-trained Markov Random Field (MRF) model for multi-scale estimation. Liu~\etal~\cite{liu2014discrete} built a discrete and continuous Conditional Random Field (CRF) model for fusing both local and global features. The second group of the methods is based on the advanced deep learning models~\cite{eigen2015predicting,liu2015deep,wang2015towards,roymonocular,laina2016deeper}. Eigen~\etal~\cite{eigen2014depth} developed a multi-scale CNN for fusing both coarse and fine predictions from different semantic layers of the CNN. Recently, researchers studied implementing the CRF models with CNN enabling the end-to-end optimization of the whole deep network~\cite{liu2015deep,xu2017multi,xu2017learning}. 
\par Many efforts have been devoted to the scene parsing task in recent years. The scene parsing task is usually treated as a pixel-level prediction problem and the performance is greatly boosted by the fully convolutional strategy~\cite{long2015fully} which replaces the full connected layers with convolutional layers and dilated convolution~\cite{chen2016deeplab, yu2015multi}. The other works mainly focused on multi-scale feature learning and ensembling~\cite{chen2016attention, xia2016zoom, hariharan2015hypercolumns}, end-to-end structure prediction with CRF models~\cite{liu2015semantic, arnab2016higher, zheng2015conditional, xu2018structured} and designing convolutional encoder-decoder network structures~\cite{noh2015learning, badrinarayanan2015segnet}. These works focused on an individual task but not jointly optimizing the depth estimation and scene parsing together.
 \par Some works~\cite{mousavian2016joint, wang2015towards, jafari2017analyzing, kuga2017multi} explored simultaneously learning the depth estimation and the scene parsing tasks. For instance, Wang~\etal~\cite{wang2015towards} introduced an approach to model the two tasks within a hierarchical CRF, while the CRF model is not jointly learned with the CNN. However, these works directly learn the two tasks without treating them as multi-modal input for the final tasks.

\par\textbf{Deep multi-task learning for vision.} Deep multi-task learning\cite{misra2016cross,ruder2017sluice} has been widely used in various computer vision problems, such as joint inference scene geometric and semantic~\cite{kendall2017multi},  face attribute estimation~\cite{han2017heterogeneous}, simultaneous contour detection and semantic segmentation~\cite{gupta2013perceptual}. Yao and Urtasun~\etal~\cite{yao2012describing} proposed an approach for joint learning three tasks \ie object detection, scene classification and semantic segmentation. Hariharan~\etal~\cite{hariharan2014simultaneous} proposed to simultaneously learn object detection and semantic segmentation based on the R-CNN framework. However, none of them considered introducing multi-task prediction and multi-modal distillation steps at the intermediate level of a CNN to improve the target tasks.

\section{PAD-Net: Multi-Tasks Guided Prediction-and-Distillation Network}
\vspace{-4pt}
In this section, we describe the proposed PAD-Net for simultaneous depth estimation and scene parsing. We first present an overview of the proposed PAD-Net, and then introduce the details of the PAD-Net. Finally, we  illustrate the optimization and inference schemes for the overall network.
\vspace{-15pt}
\subsection{Approach Overview}
\vspace{-3pt}

\par Figure~\ref{framework} depicts the framework of the proposed multi-tasks guided prediction and distillation network (PAD-Net). PAD-Net consists of four main components. First, a front-end fully convolutional encoder produces deep features. Second, an intermediate multi-task prediction module, which uses the deep features in the previous component for generating intermediate predictions.
Third, a multi-modal distillation module which is used for incorporating useful multi-modal information from the intermediate predictions to improve the final tasks. 
Fourth, the decoders uses the distilled information for depth estimation and scene parsing. 
The input of PAD-Net is RGB images during both training and testing, and the final output is the depth and semantic parsing maps. During training, ground-truth labels for scene parsing, depth estimation and other two intermediate tasks, \ie~surface normal estimation and contour prediction, are used. Although four different kinds of supervision are used, we do not require extra annotation effort, since the surface normal and the contours can be directly inferred from depth and semantic labels, respectively.

\begin{figure*}[!t]
\centering
\includegraphics[width=6.75in]{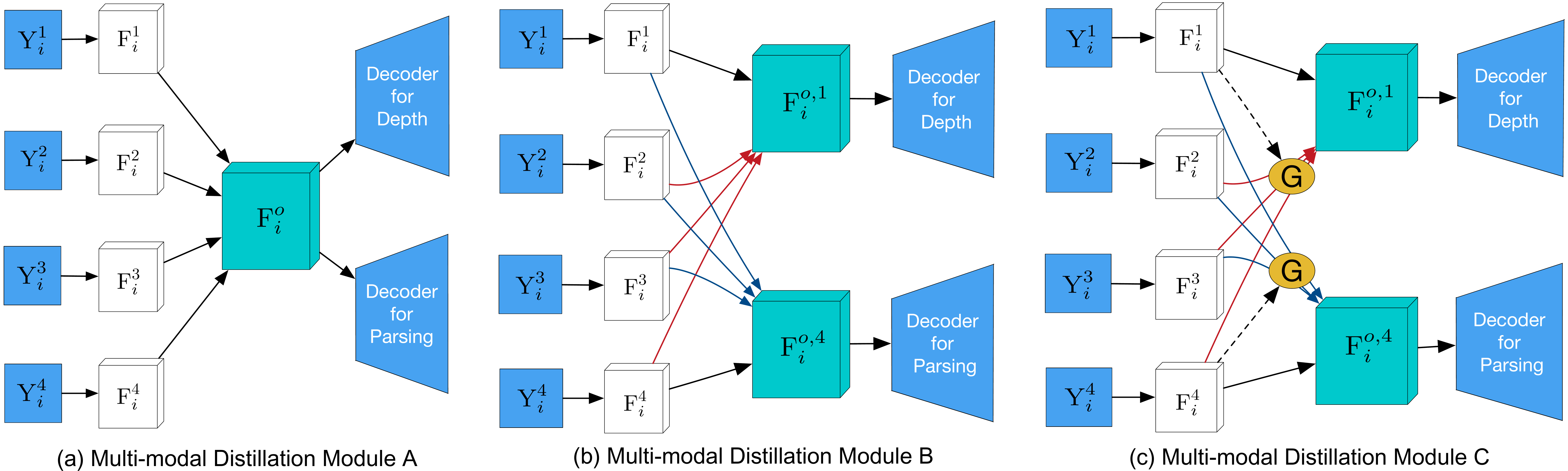} 
\caption{Illustration of the designed different multi-modal distillation modules. The symbols $\textsc{Y}_i^1$, $\textsc{Y}_i^2$, $\textsc{Y}_i^3$, $\textsc{Y}_i^4$ represent the predictions corresponding to multiple intermediate tasks. The distillation module A is a naive combination of the multiple predictions; the module B proposes a mechanism of passing message between different predictions; the module C shows an attention-guided message passing mechanism for distillation. The symbol $\textsc{G}$ denotes a generated attention map which is used as guidance in the distillation. }
\label{distillation}
\vspace{-12pt}
\end{figure*}
\vspace{-3pt}
\subsection{Front-End Network Structure}
\vspace{-3pt}
The front-end backbone CNN could employ any network structures, such as the commonly used AlexNet~\cite{krizhevsky2012imagenet}, VGG~\cite{simonyan2014very} and ResNet~\cite{he2015deep}. To obtain better deep representations for predicting multiple intermediate tasks, we do not directly use the features from the last convolutional layer of the backbone CNN. A multi-scale feature aggregation procedure is performed to enhance the last-scale feature map via combining the previous scales feature maps derived from different semantic layers of the backbone CNN, as shown in Figure~\ref{framework}. The larger-resolution feature maps from shallower layers are down-sampled via convolution and bilinear interpolation operations to the resolution of the last-scale feature map. The convolution operations are also used to control the number of feature channels to make the feature aggregation more memory efficient. And then all the re-scaled feature maps are concatenated for the follow up deconvolutional operations. Similar to~\cite{chen2014semantic, yu2015multi}, we also apply the dilated convolution strategy in the front-end network to produce feature maps with enlarged receptive field. 

\vspace{-3pt}
\subsection{Deep Multi-Task Prediction}
\vspace{-3pt}
Using deep features from the front-end CNN, we perform deconvolutional operations to generate four sets of task-specific feature maps. We obtain features with $N$ channels for the main depth estimation and scene parsing tasks while features with $N/2$ channels for the other two auxiliary tasks. The feature map resolution is made to be the same for four tasks and to be $2\times$ as that of the front-end feature maps. Then separate convolutional operations are performed to produce the score maps for the corresponding four tasks. The score maps are made to be $1/4$ as the resolution of the input RGB images via the bilinear interpolation. Four different loss functions are added for learning the four intermediate tasks with the re-scaled ground-truth maps. It should be noted that the intermediate multi-task learning not only provides deep supervision for optimizing the front-end CNN, but also helps to provide valuable multi-modal predictions, which are further used as input for the final tasks.
\subsection{Deep Multi-Modal Distillation}
\vspace{-3pt}

As mentioned before, the deep multi-modal distillation module fuses information from the intermediate predictions for each specific final task. It aims at effectively utilizing the complementary information from the intermediate predictions of relevant tasks. To achieve this target and under our general framework, it is potentially flexible to use any distillation scheme. In this paper, we develop and investigate three different module designs as shown in Figure~\ref{distillation} to show how the multi-modal distillation helps improving the final tasks. 
The distillation module A represents a naive concatenation of the features extracted from these predictions. The distillation module B passes message between different predictions. The distillation module C is an attention-guided message passing mechanism for information fusion. To generate richer information and bridge the gap between these predictions, before the distillation procedure, all the intermediate prediction maps associated with the $i$-th training sample, denoted by $\{\mathrm{Y}_i^t\}_{t=1}^{T}$, are first correspondingly transformed to feature maps $\{\mathrm{F}_i^t\}_{t=1}^{T}$ with more channels via convolutional layers, where $T$ is the number of intermediate tasks.
\par\textbf{Multi-Modal Distillation module A.} A common way in deep networks for information fusion is to perform a naive concatenation of the feature maps or the score maps from different semantic layers of the network. We aslo consider this simple scheme as our basic distillation module. The module A outputs only one set of fused feature maps via $\mathrm{F}_i^o \leftarrow \textsc{CONCAT}(\mathrm{F}_i^1, ..., \mathrm{F}_i^{T})$, where $\textsc{CONCAT}(\cdot)$ denotes the concatenation operation. And then $\mathrm{F}_i^o$ is fed into different decoders for predicting different final tasks, \ie~the depth estimation and the scene parsing tasks.
\par\textbf{Multi-Modal Distillation module B.} The module A outputs the same set of feature maps for the two final tasks. Differently, the module B learns a separate set of feature maps for each final task. For the $k$-th final task, let us denote $\mathrm{F}_i^{k}$ as the feature maps before message passing and denote $\mathrm{F}_i^{o, k}$ as the feature maps after the distillation. We refine $\mathrm{F}_i^{k}$ via passing message from the feature maps of other tasks as follows:
 \begin{equation}
\setlength{\abovedisplayskip}{1pt}
 \mathrm{F}_i^{o, k} \leftarrow \mathrm{F}_i^{k} + \sum_{t=1(\neq k)}^{T} (\mathrm{W}_{t, k} \otimes \mathrm{F}_i^{t}),
 \label{eq:message1}
\setlength{\belowdisplayskip}{1pt}
 \end{equation}
where $\otimes$ denotes convolution operation, and $\mathrm{W}_{t, k}$ denotes the parameters of the convolution kernel corresponding to the $t$-th feature map and the $k$-th feature map. 
Then the obtained feature map $\mathrm{F}_i^{o, k}$ is used by the decoded for the corresponding $k$-th task. By using the task-specific distillation feature maps, the network can preserve more information for each individual task and is able to facilitate smooth convergence.
\vspace{-2pt}
\par\textbf{Multi-Modal Distillation module C.} The module C introduces an attention mechanism for the distillation task. The attention mechanism~\cite{mnih2014recurrent} has been successfully applied in various tasks such as image caption generation~\cite{xu2015show} and machine translation~\cite{luong2015effective} for selecting useful information. Specifically, we utilize the attention mechanism for guiding the message passing between the feature maps generated from different madalities for different tasks. Since the passed information flow is not always useful, the attention can act as a gate function to control the flow, in other words to make the network automatically learn to focus or to ignore information from other features. When we pass message to the $k$-th task, an attention map $\mathrm{G}_i^{k}$ is first produced from the corresponding set of feature maps $\mathrm{F}_i^{k}$ as follows:
 \begin{equation}
 \setlength{\abovedisplayskip}{1pt}
 \mathrm{G}_i^{k} \leftarrow \sigma({\mathrm{W}_g^{k} \otimes \mathrm{F}_i^{k}}),
 \label{eq:message1}
 \setlength{\belowdisplayskip}{1pt}
 \end{equation}
 where $\mathrm{W}_g^{k}$ is the convolution parameter and $\sigma$ is a sigmoid function for normalizing the attention map. Then the message is passed with the attention map controlled as follows:
  \begin{equation}
  \setlength{\abovedisplayskip}{1pt}
 \mathrm{F}_i^{o, k} \leftarrow \mathrm{F}_i^{k} + \sum_{t=1(\neq k)}^{T}  \mathrm{G}_i^{k} \odot (\mathrm{W}_t \otimes \mathrm{F}_i^{t}),
 \label{eq:message2}
 \setlength{\belowdisplayskip}{1pt}
 \end{equation}
  where $\odot$ denotes element-wise multiplication. 

\subsection{Decoder Network Structure}
For the task-specific decoders, we use two consecutive deconvolutional layers to up-sample the distilled feature maps for pixel-level prediction. Since the distilled feature maps have a resolution of $1/4$ to that of the input RGB image, each deconvolutional layer 2 time up-scales in resolution and accordingly reduces the number of output channels by half. Finally we use a convolution operation to generate the score maps for each final task.

\subsection{PAD-Net Optimization}
\par\textbf{End-to-end network optimization.} We have four intermediate prediction tasks, \ie~two discrete classification problems (scene parsing and contour prediction) and two continuous regression problems (surface normal estimation and depth estimation). However, we only require the annotations of the semantic labels and the depth, since the contour labels can be generated from the semantic labels and the surface normal can be calculated from the depth map. As our final target is to simultaneously perform the depth estimation and scene parsing, the whole network needs to optimize six losses with four different types. Specifically, we use a cross-entropy loss for the contour prediction task, a softmax loss for the scene parsing task and an Euclidean loss for both the depth and surface normal estimation tasks. Since the groundtruth depth maps have invalid points, we mask these points during training. Similar to previous works~\cite{sermanet2013overfeat, teichmann2016multinet}, we jointly learn the whole network with a linearly combined optimization objective, \ie~$L_{all} = \sum_{i=1}^6 w_i * L_i$, where $L_i$ is the loss for the $i$-th task and $w_i$ is the corresponding loss weight. 
\par\textbf{Inference.} During the inference, We obtain the prediction results from the separate decoders. One important advantage of the PAD-Net is that  it is able to incorporate rich domain knowledge from different predictions, \ie~scene semantic, depth, surface normal and object contours, while it only requires a single RGB image for the inference. 


\vspace{-6pt}
\section{Experiments}
\vspace{-6pt}
To demonstrate the effectiveness of the proposed approach for simultaneous depth recovery and scene parsing, we conduct experiments on two publicly available benchmark datasets which provide both the depth and the semantic labels, including an indoor dataset NYU depth V2 (NYUD-v2)~\cite{silberman2012indoor} and an outdoor dataset Cityscapes~\cite{Cordts2016Cityscapes}. In the following we describe the details of our experimental evaluation.

\begin{figure*}[!t]
\centering
\includegraphics[width=6.8in]{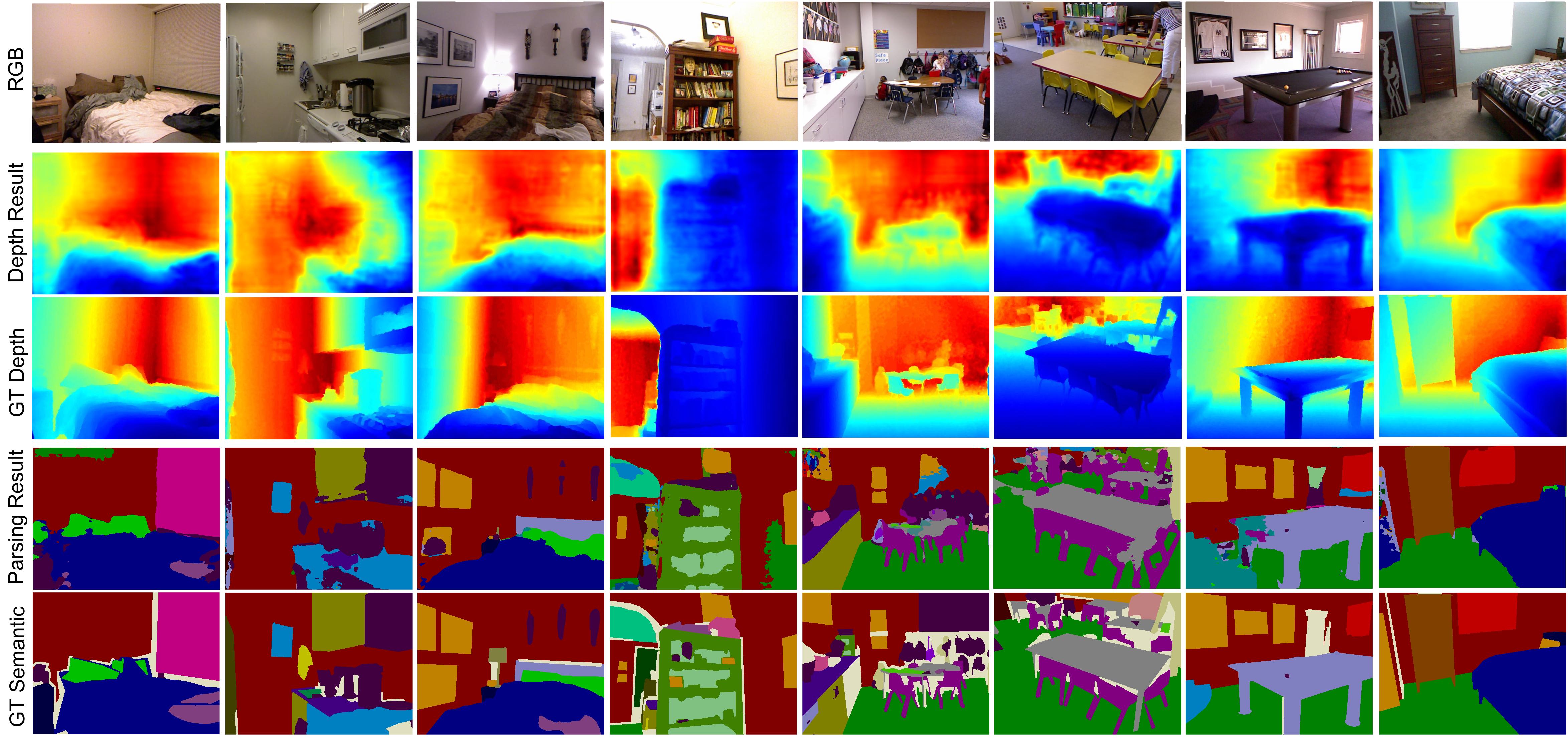} 
\caption{Quanlitative examples of depth prediction and 40-classes scene parsing results on the NYUD-v2 dataset. The second and the four row are the estimated depth maps and the scene parsing results from the proposed PAD-Net, respectively.}
\label{nyu2examples}
\vspace{-10pt}
\end{figure*}
\vspace{-3pt}
\subsection{Experimental Setup}
\vspace{-3pt}
\textbf{Datasets and Data Augmentation.}
The \textbf{NYUD-v2} dataset~\cite{silberman2012indoor}  is a popular indoor RGBD dataset, which has been widely used for depth estimation~\cite{eigen2014depth} and semantic segmentation~\cite{gupta2014learning}. It contains 1449 pairs of RGB and depth images captured from a Kinect sensor, in which 795 pairs are used for training and the rest 654 for testing. Following~\cite{gupta2014learning}, The training images are cropped to have a resolution of $560\times 425$. The training data are augmented on the fly during the training phase. The RGB and depth images are scaled with a randomly selected ratio in $\{1, 1.2, 1.5\}$ and the depth values are divided by the ratio. We also flip the training samples with a possibility of 0.5. 

The \textbf{Cityscapes}~\cite{Cordts2016Cityscapes} is a large-scale dataset mainly used for semantic urban scene understanding. The dataset is collected over 50 different cities spanning several months, and overall 19 semantic classes are annotated. The fine-annotated part consists of training, validation and test sets containing 2975, 500, and 1525 images, respectively. The dataset also provides pre-computed disparity depth maps associated with the rgb images. Similar to NYUD-v2, we perform the data augmentation on the fly by scaling the images with a selected ratio in $\{0.5, 0.75, 1, 1.25, 1.75\}$ and randomly flipping them with a possibility of 0.5. As the images of the dataset have a high resolution ($2048 \times 1024$), we crop the image with size of $640 \time 640$ for training due to the limitation of the GPU memory. 

\textbf{Evaluation Metrics.} For evaluating the performance of the depth estimation, we use several quantitative metrics following previous works~\cite{eigen2014depth, liu2015deep, xu2017multi}, including (a) mean relative error (rel): \( \frac{1}{N}\sum_p\frac{| d_p - d_p^* |}{d_p} \); (b) root mean squared error (rms): 
\( \sqrt{\frac{1}{N}\sum_{p}({d}_p - d_p^*)^2} \); 
(c) mean log10 error (log10): \( \frac{1}{N}\sum_i \Vert \log_{10}(d_p) - \log_{10}(d_p^*) \Vert \) and
(d) accuracy with threshold $t$: percentage (\%) of $d_p^*$ subject to $\max (\frac{d_p^*}{d_p}, \frac{d_p}{d_p^*}) = 
\delta < t~(t \in [1.25, 1.25^2, 1.25^3])$, where $d_p$ and $d_p^*$ are the prediction and the groundtruth depth at the $p$-th pixel, respectively. For the evaluation of the semantic segmentation,  we adopt three commonly used metrics, \ie~mean Intersection over Union (mIoU), mean accuracy and pixel accuracy. The mean IoU is calculated via averaging the Jaccard scores of all the predicted classes. The mean accuracy is the accuracy among all classes and pixel accuracy is the total accuracy of pixels regardless of the category. On the Cityscapes, both the pixel-level mIoU and instance-level mIoU are considered.

\textbf{Implementation Details. }
The proposed network structure is implemented base on \textit{Caffe} library~\cite{jia2014caffe} and on Nvidia Titan X GPUs. The front-end convolutional encoder of PAD-Net naturally supports any network structure. During the training, the front-end network is first initialized with parameters pre-trained with ImageNet for training, and the rest of the network is randomly intialized. The whole training process is performed with two phases. In the first phase, we only optimize the front-end network with the scene parsing task and use a learning rate 0.001. After that, the whole network is jointly trained with multi-task losses and a lower learning rate of 10e-5 is used for a smooth convergence. As the final tasks are depth estimation and scene parsing, we set the loss weight of the contour prediction and surface normal estimation as 0.8. In the multi-task prediction module, $N$ is set to 512. Total 60 epochs are used for NYUD-v2, and 40 epochs for Cityscapes. Due to the sparse groundtruth depth maps of the Cityscapes dataset, the invalid points are masked out in the backpropagation. The network is optimized using stachastic gradient descent with the weight decay and the momentum set to 0.0005 and 0.99, respectively.  

\begin{table}
 \centering
 \caption{Diagnostic experiments for the depth estimation task on NYUD-v2 dataset. Distillation A, B, C represents the proposed three multi-modal distillation modules.}
 \Huge
\setlength\tabcolsep{10pt}
 \renewcommand\arraystretch{1.1} 
\resizebox{1.02\linewidth}{!} {
\begin{tabular}{l|ccc|ccc}
\toprule[2pt]
\multirow{2}{*}{Method} & \multicolumn{3}{c|}{\tabincell{c}{Error (lower is better)}} & \multicolumn{3}{c}{\tabincell{c}{Accuracy (higher is better)}} \\ \cline{2-7}
                                      & rel & log10 & rms & $\delta < 1.25$ & $\delta < 1.25^2$ & $\delta < 1.25^3$ \\ \midrule
Front-end + DE (baseline)  & 0.265 & 0.120 &0.945  &0.447 & 0.745 & 0.897 \\
Front-end + DE + SP (baseline) & 0.260    & 0.117 & 0.930   &    0.467    &     0.760    &    0.905    \\
PAD-Net (Distillation A + DE) & 0.248 & 0.112 & 0.892  &    0.513     &    0.798    &   0.921     \\
PAD-Net (Distillation B + DE)  &   0.230       &  0.099    &    0.850     &    0.591    &    0.854  &  0.953 \\
PAD-Net (Distillation C + DE)   & 0.221 & 0.094 & 0.813   &  0.619 &  0.882 & 0.965 \\
PAD-Net (Distillation C + DE + SP) & \textbf{0.214} & \textbf{0.091} & \textbf{0.792} &\textbf{ 0.643} & \textbf{0.902} & \textbf{0.977} \\
\bottomrule[2pt]                           
\end{tabular}
}
\label{abalation_nyu_depth}
\vspace{-8pt}
\end{table}

\begin{table}
 \centering
 \caption{Diagnostic experiments for the scene parsing task on the NYUD-v2 dataset.}
\Huge
\setlength\tabcolsep{10pt}
 \renewcommand\arraystretch{1.1} 
\resizebox{1.02\linewidth}{!} {
\begin{tabular}{l|ccc}
\toprule[2pt]
Method              & Mean IoU  &    Mean Accuracy     &   Pixel Accuracy \\\midrule
Front-end + SP (baseline) &  0.291 &  0.301 &  0.612 \\
Front-end + SP + DE (baseline) &     0.294         &      0.312     &    0.615    \\
PAD-Net (Distillation A  + SP)   &     0.308   &    0.365     &   0.628 \\
PAD-Net (Distillation B  + SP) &    0.317       &        0.411               &    0.638 \\
PAD-Net (Distillation C  + SP)  &      0.325       &         0.432         &     0.645 \\
PAD-Net (Distillation C + DE + SP) &  \textbf{0.331}   &    \textbf{0.448}   &  \textbf{0.647} \\
\bottomrule[2pt]                       
\end{tabular}
}
\label{abalation_nyu_parsing}
\vspace{-8pt}
\end{table}

\begin{table}
 \centering
 \caption{Quantitative comparison with state-of-the-art methods methods on the scene parsing task on the NYUD-v2 dataset. The methods `Gupta~\etal'~\cite{gupta2014learning} and `Arsalan~\etal'~\cite{mousavian2016joint} jointly learn two tasks.}
\Huge
\setlength\tabcolsep{10pt}
\resizebox{1.02\linewidth}{!} {
\begin{tabular}{l|c|ccc}
\toprule[2pt]
Method          &    Input Data Type    & Mean IoU  &    Mean Accuracy     &   Pixel Accuracy \\\midrule
Deng~\etal~\cite{deng2015semantic} &     RGB + Depth                 &    -          &      0.315                    &    0.638    \\
FCN~\cite{long2015fully}   &  RGB                 &     0.292        &       0.422                &   0.600 \\
FCN-HHA~\cite{long2015fully} &   RGB + Depth         &     0.340        &        0.461               &    0.654 \\
Eigen and Fergus~\cite{eigen2015predicting}  & RGB           &      0.341       &         0.451              &     0.656 \\
Context~\cite{lin2016efficient} & RGB  & 0.406 & 0.536 & 0.700 \\
Kong~\etal~\cite{kong2017recurrent} & RGB & 0.445 & - & 0.721 \\ 
RefineNet-Res152~\cite{lin2016refinenet} & RGB  & 0.465 & 0.589 & 0.736 \\\midrule
Gupta~\etal~\cite{gupta2014learning} & RGB + Depth & 0.286 & - & 0.603 \\
Arsalan~\etal~\cite{mousavian2016joint} &  RGB   &        0.392              &      0.523     &      0.686 \\
\midrule
PAD-Net-ResNet50 (Ours) &  RGB   &        \textbf{0.502}              &      \textbf{0.623}     &      \textbf{0.752} \\
\bottomrule[2pt]                       
\end{tabular}
}
\label{sota_nyu_parsing}
\vspace{-8pt}
\end{table}


\subsection{Diagnostics Experiments}
To deeply analyze the proposed approach and demonstrate its effectiveness, we conduct diagnostics experiments on both NYUD-v2 and Cityscapes datasets. For the front-end network, according to the complexity of the dataset, we choose AlexNet~\cite{krizhevsky2012imagenet} and ResNet-50~\cite{he2015deep} network structures for NYUD-v2 and Cityscapes, respectively.

\par\textbf{Baseline methods and different variants of PAD-Net.} To conduct the diagnostic experiments, we consider two baseline methods and different variants of the proposed PAD-Net. The baseline methods include: (i) Front-end + DE: performing the depth estimation (DE) task with the front-end CNN; (ii) Front-end + SP + DE: performing the scene parsing (SP) and the depth estimation tasks simultaneously with the front-end CNN. The different variants include: (i) PAD-Net (Distillation A + DE): PAD-Net performing the DE task using the distillation module A; (ii) PAD-Net (Distillation B + DE): similar to (i) while using the distillation module B; (iii) PAD-Net (Distillation B + DE): similar to (i) while using the distillation module C; (iv) PAD-Net (Distillation C + DE + SP): performing DE and SP tasks simultaneously with the distillation module C.

\begin{table}
 \centering
 \caption{Quantitative comparison with state-of-the-art methods on the depth estimation task on NYUD-v2 dataset. The methods `Joint HCRF'~\cite{wang2015towards} and `Jafari~\etal'~\cite{jafari2017analyzing} simultaneously learn the two tasks.}
\setlength\tabcolsep{8pt}
\resizebox{1.01\linewidth}{!} {
\begin{tabular}{l|c|ccc|ccc}
\toprule
\multirow{2}{*}{Method} & \multirow{2}{*}{\# of Training} & \multicolumn{3}{c|}{\tabincell{c}{Error (lower is better)}} & \multicolumn{3}{c}{\tabincell{c}{Accuracy (higher is better)}} \\ \cline{3-8}
                                 &     & rel & log10 & rms & $\delta < 1.25$ & $\delta < 1.25^2$ & $\delta < 1.25^3$ \\ \midrule
Saxena \etal~\cite{saxena2009make3d}     &    795            & 0.349 &  -     & 1.214  &0.447 & 0.745 & 0.897 \\
Karsch \etal~\cite{karsch2014depth}           &  795  &0.35    &0.131&1.20   &    -     &     -    &    -      \\
Liu \etal~\cite{liu2014discrete} &  795 & 0.335 & 0.127 & 1.06  &    -     &     -    &     -     \\
Ladicky \etal~\cite{ladicky2014pulling}    &     795      &   -       &    -    &    -     & 0.542& 0.829&  0.941 \\
Zhuo \etal~\cite{zhuo2015indoor}         &  795            & 0.305 &0.122& 1.04  & 0.525& 0.838& 0.962 \\
Liu \etal~\cite{liu2015deep}         &             795           & 0.230 &0.095& 0.824& 0.614& 0.883&  0.975 \\
Eigen \etal~\cite{eigen2014depth}  &        120K            & 0.215  &  -      & 0.907& 0.611 &  0.887  &  0.971\\
Roi~\etal~\cite{roymonocular}         &          795                & 0.187  &  0.078& 0.744& - &  -  &  -\\
Eigen and Fergus~\cite{eigen2015predicting}   &    795        & 0.158  & -       & 0.641& 0.769 & 0.950 & 0.988 \\
Laina \etal~\cite{laina2016deeper}         &     96K       & 0.129  &0.056& {0.583}&0.801 & 0.950 & 0.986\\
Li \etal \cite{li2017monocular}  & 96K & 0.139 & 0.058 & 0.505 & 0.820   & 0.960 &   0.989 \\
Xu \etal \cite{xu2017multi}         &         4.7K             & 0.139  & 0.063 & 0.609 &  0.793 & 0.948 & 0.984 \\
Xu \etal \cite{xu2017multi}         &         95K             & 0.121  & \textbf{0.052} & 0.586 &  0.811 & 0.950 & 0.986 \\ \midrule
Joint HCRF~\cite{wang2015towards}     &     795         & 0.220 & 0.094&0.745& 0.605 & 0.890 & 0.970 \\
Jafari~\etal~\cite{jafari2017analyzing} & 795 & 0.157 & 0.068 & 0.673 & 0.762 & 0.948 & 0.988 \\\midrule
PAD-Net-ResNet50 (Ours)     &  795  & \textbf{0.120}  & 0.055 & \textbf{0.582} & \textbf{0.817}  & \textbf{0.954 }& \textbf{0.987} \\
\bottomrule                           
\end{tabular}
}
\label{sota_nyu_depth}
\vspace{-5pt}
\end{table}

\begin{table}[!t]
 \centering
 \caption{Quantitative comparison results with the state-of-the-art methods on the Cityscapes \textit{test} set. Our model is trained only on the fine-annotation dataset.}
\Large
\setlength\tabcolsep{10pt}
\resizebox{0.92\linewidth}{!} {
\begin{tabular}{l|cccc}
\toprule[1pt]
Method              & IoU cla.  &   iIoU cla.   &   IoU cat. & iIoU cat. \\\midrule
SegNet~\cite{badrinarayanan2015segnet} & 0.561 & 0.342 & 0.798 & 0.664  \\
CRF-RNN~\cite{zheng2015conditional} &  0.625  &  0.344  &  0.827 & 0.660  \\
SiCNN~\cite{krevso2016convolutional}   &   0.663    &  0.449  &  0.850  & 0.712  \\
DPN~\cite{liu2015semantic}  &    0.668   &  0.391  &   0.860  &  0.691 \\
Dilation10~\cite{yu2015multi}  &  0.671 &  0.420 &  0.865   &  0.711  \\
LRR~\cite{ghiasi2016laplacian} &   0.697   &   0.480  &   0.882 & 0.747  \\
DeepLab~\cite{chen2016deeplab} &    0.704  &   0.426   &   0.864 & 0.677 \\
Piecewise~\cite{lin2016efficient} &  0.716   &   0.517     &   0.873 & 0.741\\
PSPNet~\cite{zhao2016pyramid}  &  0.784     &  0.567     &   0.906 & 0.786 \\\midrule
PAD-Net-ResNet101 (Ours) &  0.803     &  0.588     &   0.908 & 0.785 \\
\bottomrule[1pt]                       
\end{tabular}
}
\label{sota_cityscapes}
\vspace{-5pt}
\end{table}

\begin{table}[!h]
 \centering
\caption{Quantitative evaluation of the importance of intermediate supervision and multiple tasks.}
 \Huge
\setlength\tabcolsep{10pt}
 \renewcommand\arraystretch{1.1} 
\resizebox{0.85\linewidth}{!} {
\begin{tabular}{l|ccc|ccc}
\toprule[2pt]
\multirow{2}{*}{Method} & \multicolumn{3}{c|}{\tabincell{c}{Depth Metrics}} & \multicolumn{3}{c}{\tabincell{c}{Parsing Metrics}} \\ \cline{2-7}
                                      & rel & log10 & rms & Mean IoU & Mean Acc & Pixel Acc \\ \midrule
MTDN-mds  & 0.149 & 0.063 & 0.701  & 0.474 & 0.597 & 0.727 \\\hline
MTDN-inp0 & 0.153  & 0.069 & 0.721  & 0.465 & 0.588 & 0.713  \\
MTDN-inp2  & 0.139 & 0.064 & 0.672  & 0.481 & 0.603 & 0.729   \\
MTDN-inp3  & 0.128 & 0.059 & 0.617  & 0.490 & 0.612 & 0.739   \\\midrule
MTDN-full & \textbf{0.120} & \textbf{0.055} & \textbf{0.582} &\textbf{ 0.502} & \textbf{0.623} & \textbf{0.752} \\
\bottomrule[2pt]                           
\end{tabular}
}
\label{abalation_supervision_tasks}
\vspace{-10pt}
\end{table}

\par
\begin{figure}[!t]
\centering
\includegraphics[width=3.3in]{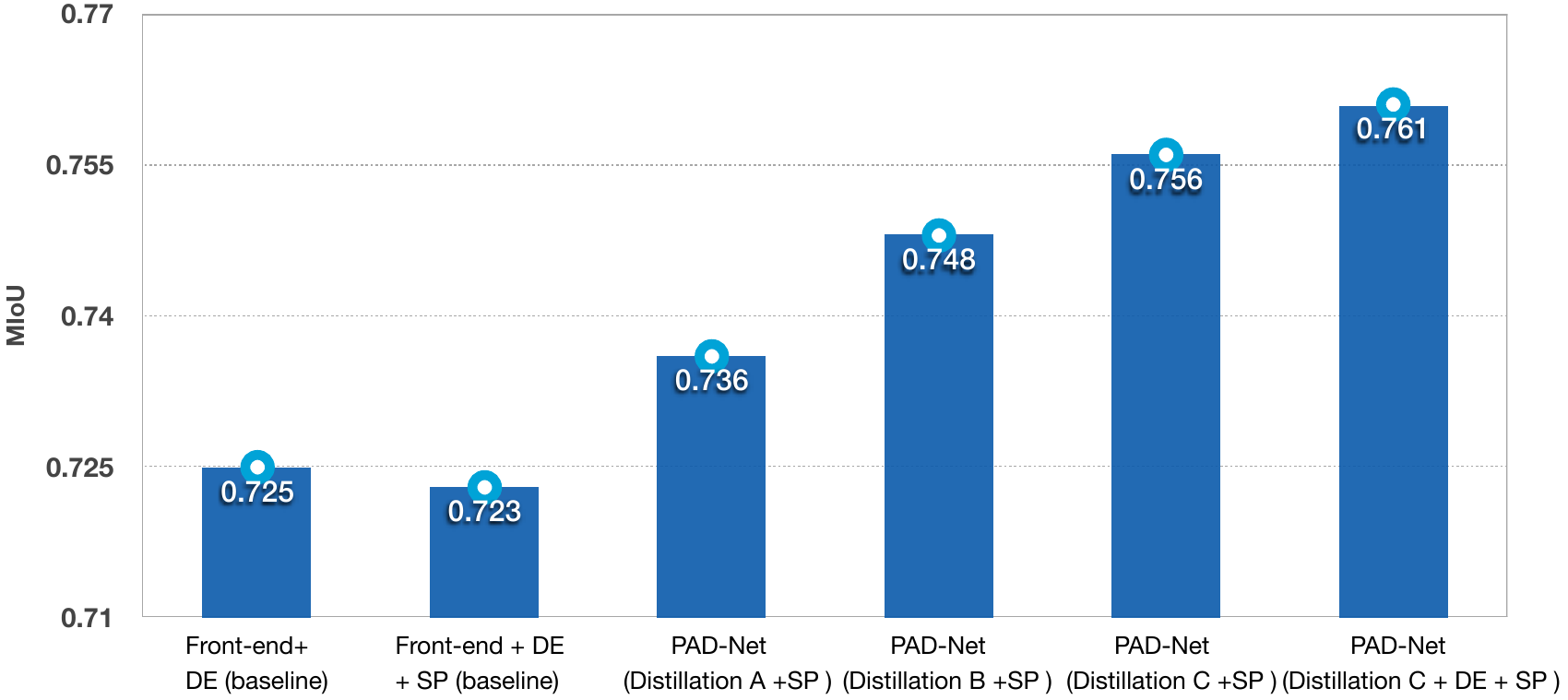} 
\caption{Diagnostic experiments of the proposed approach for the scene parsing on Cityscapes \textit{val} dataset with ResNet-50 as the front-end backbone CNN.}
\label{cityscapes_ablation_study}
\vspace{-5pt}
\end{figure}

\textbf{Effect of direct multi-task learning.} To investigate the effect of simultaneously optimizing two different task as previous works~\cite{mousavian2016joint, wang2015towards}, \ie predicting two different tasks directly from the last scale feature map of the front-end CNN. We carry out experiments on both the NYUD-v2 and Cityscapes datasets, as shown in Table.~\ref{abalation_nyu_depth}, ~\ref{abalation_nyu_parsing} and Figure~\ref{cityscapes_ablation_study}, respectively. It can be observed that on NYUD-v2, the Front-end + DE + SP slightly outperforms the Front-end + DE, while on Cityscapes, the performance of Front-end + DE + SP is even decreased, which means that using a direct multi-task learning as traditional is probably not an effective means to facilitate each other the performance of different tasks.
\par\textbf{Effect of multi-modal distillation.} We further evaluate the effect of the proposed three different distillation modules for incorporating information from different prediction tasks. Table~\ref{abalation_nyu_depth} shows the results on the depth prediction task using PAD-Net embedded with the distillation module A, B and C. It can be seen that these three variants of PAD-Net are all obviously better than the two baseline methods, and the best one of ours, PAD-Net (Distillation C + DE) is 4.4 and 2.3 points better than the baseline Front-end + DE + SP on the rel and on the log10 metric respectively, and on the segmentation task on the same dataset, it is 3.1 points higher than the same baseline on the mIoU metric, which clearly demonstrates the effectiveness of the proposed multi-modal distillation strategy. Similar performance gaps can be also observed on the segmentation task on Cityscapes in Figure~\ref{cityscapes_ablation_study}. For comparing the different distillation modules, the message passing between different tasks (the module B and C) significantly boosts the the performance over the naive combination method (the module C). By using the attention guided scheme, the performance of the module C is further improved over the module B.

\begin{figure*}[!t]
\centering
\includegraphics[width=6.8in]{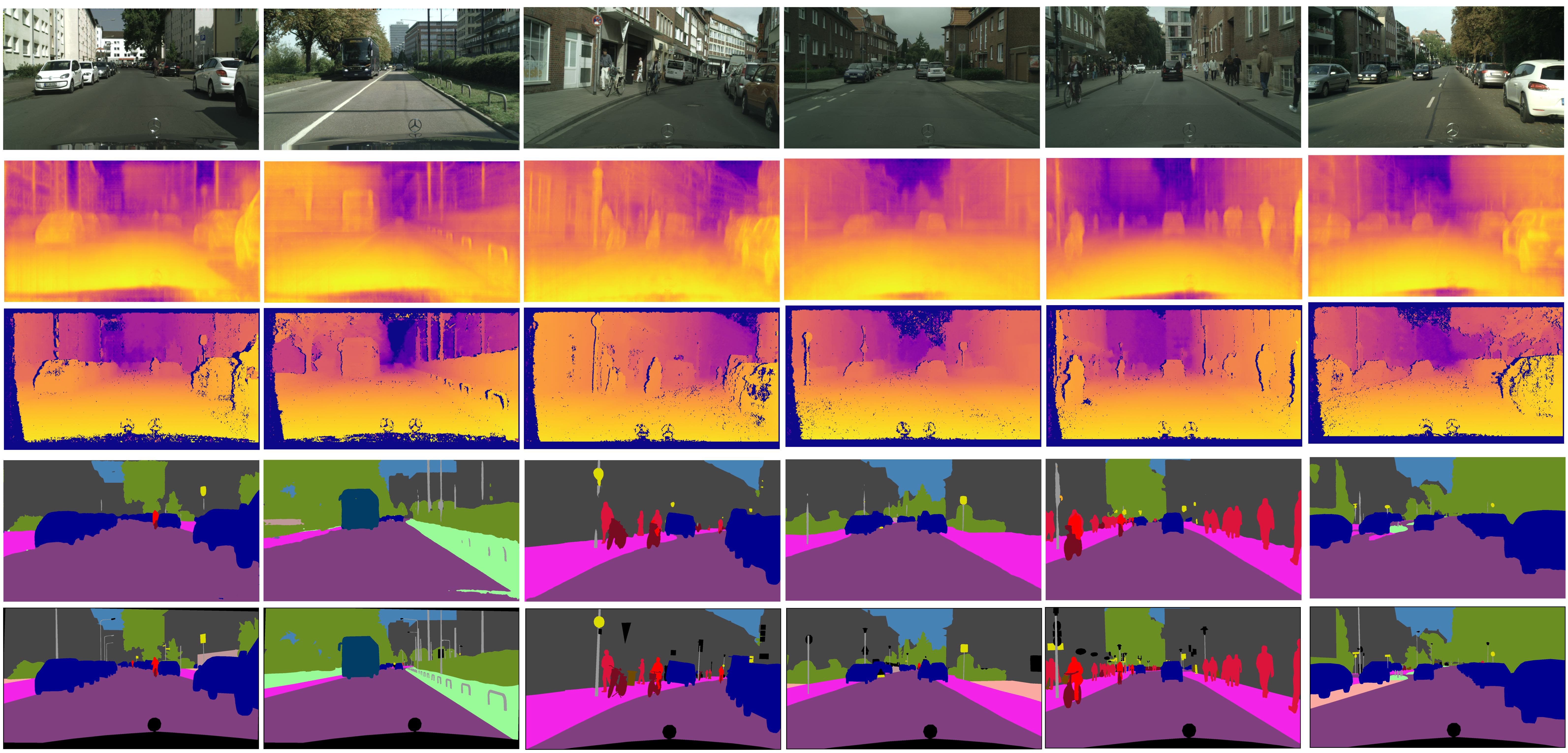} 
\caption{Quanlitative examples of depth prediction and 19-classes scene parsing results the Cityscapes dataset. The second and the fourth row correspond to the sparse depth and the semantic groundtruth, respectively.}
\label{cityscapesexamples}
\vspace{-10pt}
\end{figure*}

\par\textbf{Effect of multi-task guided simultaneous prediction.} We finally verify that the proposed multi-tasks guided prediction and distillation approach facilitates boosting the performance of both the depth estimation and scene parsing. The results of PAD-Net (Distillation C + DE + SP) clearly outperforms PAD-Net (Distillation C + DE) and PAD-Net (Distillation C + SP) in both the depth estimation task (Table~\ref{abalation_nyu_depth}) and the segmentation task (Tabel~\ref{abalation_nyu_parsing} and Figure~\ref{cityscapes_ablation_study}). This shows that our design of PAD-Net can use multiple final tasks in learning more effective features. More importantly, PAD-Net (Distillation C + DE + SP) obtains remarkably better performance than the baseline Front-end + DE + SP, further demonstrating the superiority of the proposed PAD-Net compared with the methods directly using two tasks to learn a deep network.

\par\textbf{Importance of intermediate supervision and tasks.} To evaluate the importance of the intermediate tasks, we use the multiple deep supervision, but consider different number of intermediate predictions for the distillation module, including MTDN-inp2 (2 inputs, depth + semantic map), MTDN-inp3 (3 inputs, depth + semantic map + surface normal) and MTDN-full (4 inputs). As shown in Table~\ref{abalation_supervision_tasks}, MTDN-mds is obviously worse than MTDN-full, meaning that the performance gain is not only because of the model capacity. MTDN-inp0 is also worse than MTDN-full, showing that the improvement is not just because of adding the intermediate supervision. To evaluate the importance of the intermediate tasks, we use the multiple deep supervision, but consider different number of intermediate predictions for the distillation module, including MTDN-inp2 (2 inputs, depth + semantic map), MTDN-inp3 (3 inputs, depth + semantic map + surface normal) and MTDN-full (4 inputs). The results  on NYUD-v2 are shown in Table 1. It is obvious that MTDN-mds is significantly worse than MTDN-full on both tasks (2.9\% worse on rel, 2.8\% worse on mIoU); Using more predictions as input gradually boosts the final performance: MTDN-full is 3.3\% (on rel) and 3.7\% (on mIoU) better than MTDN-inp0. 

\subsection{State-of-the-art Comparison}
\par\textbf{Depth estimation.} On the depth estimation task, we compare with several state-of-the-art methods, including: methods adopting hand-crafted features and deep representations~\cite{saxena2009make3d, saxena2009make3d, karsch2014depth, ladicky2014pulling, eigen2014depth, eigen2015predicting, li2017monocular, laina2016deeper}, and methods considering graphical modeling with CNN~\cite{liu2014discrete, liu2015deep, zhuo2015indoor, wang2015towards, xu2017multi}. As shown in Table~\ref{sota_nyu_depth}, PAD-Net using ResNet-50 network as the front-end achieves the best performance in all the measure metrics among all the comparison methods. It should be noted that our approach is trained only on the official training set with 795 images without using extra training data. More importantly, to compare with the methods working on joint learning the two tasks (Joint HCRF~\cite{wang2015towards}  and Jafari~\etal~\cite{jafari2017analyzing}), our performance is remarkably higher than theirs, further verifying the advantage of the proposed approach. As the Cityscapes dataset only provides the disparity map, we do not quantitatively evaluate the depth estimation performance on this dataset. Figure~\ref{nyu2examples} and~\ref{cityscapesexamples} show qualitative examples of the depth estimation on the two datasets.
\par\textbf{Scene parsing.} For the scene parsing task, we quantitatively compare the performance with the state of the art methods both on NYUD-v2 in Table~\ref{sota_nyu_parsing} and on Cityscapes in Table~\ref{sota_cityscapes}. On NYUD-v2, our PAD-Net-ResNet50 significantly outperforms the runner up competitor RefineNet-Res152~\cite{lin2016refinenet} with a 3.7 points gap on the mIoU metric. On the cityscapes, we train ours only on the fine-annotation training set, ours achieves a class-level mIoU of  0.803, which is 1.9 points better than the best competitor PSPNet trained on the same set. Qualitative scene parsing examples are shown in Figure~\ref{nyu2examples} and~\ref{cityscapesexamples}.
\vspace{-4pt}
\section{Conclusion}
\vspace{-3pt}
We have presented the proposed PAD-Net for simultaneous depth estimation and scene parsing. The PAD-Net introduces a novel deep multi-task learning means, which first predicts several intermediate auxiliary tasks and then employs the multi-task predictions as guidance to facilitate optimizing the final main tasks. Three different multi-modal distillation modules are developed to utilize the multi-task predictions more effectively. Our extensive experiments on NYUD-v2 and Cityscapes datasets demonstrated its effectiveness. We also provided new state of the art results on both the depth estimation and scene parsing tasks on NYUD-v2, and top performance on Cityscapes scene parsing task. 
\vspace{-9pt}
\section*{Acknowledgements}
\vspace{-5pt}
Wanli Ouyang is partially supported by SenseTime Group Limited. The authors would like to thank NVIDIA for GPU donation. 

{\small
\bibliographystyle{ieee}
\bibliography{egbib}

\begin{thebibliography}{10}\itemsep=-1pt

\bibitem{arnab2016higher}
A.~Arnab, S.~Jayasumana, S.~Zheng, and P.~H. Torr.
\newblock Higher order conditional random fields in deep neural networks.
\newblock In {\em ECCV}, 2016.

\bibitem{badrinarayanan2015segnet}
V.~Badrinarayanan, A.~Handa, and R.~Cipolla.
\newblock Segnet: A deep convolutional encoder-decoder architecture for robust
  semantic pixel-wise labelling.
\newblock {\em arXiv preprint arXiv:1505.07293}, 2015.

\bibitem{chen2014semantic}
L.-C. Chen, G.~Papandreou, I.~Kokkinos, K.~Murphy, and A.~L. Yuille.
\newblock Semantic image segmentation with deep convolutional nets and fully
  connected crfs.
\newblock In {\em ICLR}, 2015.

\bibitem{chen2016deeplab}
L.-C. Chen, G.~Papandreou, I.~Kokkinos, K.~Murphy, and A.~L. Yuille.
\newblock Deeplab: Semantic image segmentation with deep convolutional nets,
  atrous convolution, and fully connected crfs.
\newblock {\em arXiv preprint arXiv:1606.00915}, 2016.

\bibitem{chen2016attention}
L.-C. Chen, Y.~Yang, J.~Wang, W.~Xu, and A.~L. Yuille.
\newblock Attention to scale: Scale-aware semantic image segmentation.
\newblock In {\em CVPR}, 2016.

\bibitem{Cordts2016Cityscapes}
M.~Cordts, M.~Omran, S.~Ramos, T.~Rehfeld, M.~Enzweiler, R.~Benenson,
  U.~Franke, S.~Roth, and B.~Schiele.
\newblock The cityscapes dataset for semantic urban scene understanding.
\newblock In {\em CVPR}, 2016.

\bibitem{delage2006dynamic}
E.~Delage, H.~Lee, and A.~Y. Ng.
\newblock A dynamic bayesian network model for autonomous 3d reconstruction
  from a single indoor image.
\newblock In {\em CVPR}, 2006.

\bibitem{deng2015semantic}
Z.~Deng, S.~Todorovic, and L.~Jan~Latecki.
\newblock Semantic segmentation of rgbd images with mutex constraints.
\newblock In {\em ICCV}, 2015.

\bibitem{eigen2015predicting}
D.~Eigen and R.~Fergus.
\newblock Predicting depth, surface normals and semantic labels with a common
  multi-scale convolutional architecture.
\newblock In {\em ICCV}, 2015.

\bibitem{eigen2014depth}
D.~Eigen, C.~Puhrsch, and R.~Fergus.
\newblock Depth map prediction from a single image using a multi-scale deep
  network.
\newblock In {\em NIPS}, 2014.

\bibitem{ghiasi2016laplacian}
G.~Ghiasi and C.~C. Fowlkes.
\newblock Laplacian pyramid reconstruction and refinement for semantic
  segmentation.
\newblock In {\em ECCV}, 2016.

\bibitem{gupta2013perceptual}
S.~Gupta, P.~Arbelaez, and J.~Malik.
\newblock Perceptual organization and recognition of indoor scenes from rgb-d
  images.
\newblock In {\em CVPR}, 2013.

\bibitem{gupta2014learning}
S.~Gupta, R.~Girshick, P.~Arbel{\'a}ez, and J.~Malik.
\newblock Learning rich features from rgb-d images for object detection and
  segmentation.
\newblock In {\em ECCV}, 2014.

\bibitem{han2017heterogeneous}
H.~Han, A.~K. Jain, S.~Shan, and X.~Chen.
\newblock Heterogeneous face attribute estimation: A deep multi-task learning
  approach.
\newblock {\em arXiv preprint arXiv:1706.00906}, 2017.

\bibitem{hariharan2014simultaneous}
B.~Hariharan, P.~Arbel{\'a}ez, R.~Girshick, and J.~Malik.
\newblock Simultaneous detection and segmentation.
\newblock In {\em ECCV}, 2014.

\bibitem{hariharan2015hypercolumns}
B.~Hariharan, P.~Arbel{\'a}ez, R.~Girshick, and J.~Malik.
\newblock Hypercolumns for object segmentation and fine-grained localization.
\newblock In {\em CVPR}, 2015.

\bibitem{he2015deep}
K.~He, X.~Zhang, S.~Ren, and J.~Sun.
\newblock Deep residual learning for image recognition.
\newblock {\em arXiv preprint arXiv:1512.03385}, 2015.

\bibitem{jafari2017analyzing}
O.~H. Jafari, O.~Groth, A.~Kirillov, M.~Y. Yang, and C.~Rother.
\newblock Analyzing modular cnn architectures for joint depth prediction and
  semantic segmentation.
\newblock {\em arXiv preprint arXiv:1702.08009}, 2017.

\bibitem{jia2014caffe}
Y.~Jia, E.~Shelhamer, J.~Donahue, S.~Karayev, J.~Long, R.~Girshick,
  S.~Guadarrama, and T.~Darrell.
\newblock Caffe: Convolutional architecture for fast feature embedding.
\newblock {\em arXiv preprint arXiv:1408.5093}, 2014.

\bibitem{karsch2014depth}
K.~Karsch, C.~Liu, and S.~B. Kang.
\newblock Depth transfer: Depth extraction from video using non-parametric
  sampling.
\newblock {\em TPAMI}, 36(11):2144--2158, 2014.

\bibitem{kendall2017multi}
A.~Kendall, Y.~Gal, and R.~Cipolla.
\newblock Multi-task learning using uncertainty to weigh losses for scene
  geometry and semantics.
\newblock {\em arXiv preprint arXiv:1705.07115}, 2017.

\bibitem{kokkinos2017ubernet}
I.~Kokkinos.
\newblock Ubernet: Training a universal convolutional neural network for low-,
  mid-, and high-level vision using diverse datasets and limited memory.
\newblock In {\em CVPR}, 2017.

\bibitem{kong2017recurrent}
S.~Kong and C.~Fowlkes.
\newblock Recurrent scene parsing with perspective understanding in the loop.
\newblock {\em arXiv preprint arXiv:1705.07238}, 2017.

\bibitem{krevso2016convolutional}
I.~Kre{\v{s}}o, D.~{\v{C}}au{\v{s}}evi{\'c}, J.~Krapac, and
  S.~{\v{S}}egvi{\'c}.
\newblock Convolutional scale invariance for semantic segmentation.
\newblock In {\em GCPR}. Springer, 2016.

\bibitem{krizhevsky2012imagenet}
A.~Krizhevsky, I.~Sutskever, and G.~E. Hinton.
\newblock Imagenet classification with deep convolutional neural networks.
\newblock In {\em NIPS}, 2012.

\bibitem{kuga2017multi}
R.~Kuga, A.~Kanezaki, M.~Samejima, Y.~Sugano, and Y.~Matsushita.
\newblock Multi-task learning using multi-modal encoder-decoder networks with
  shared skip connections.
\newblock In {\em ICCVW}, 2017.

\bibitem{ladicky2014pulling}
L.~Ladicky, J.~Shi, and M.~Pollefeys.
\newblock Pulling things out of perspective.
\newblock In {\em CVPR}, 2014.

\bibitem{laina2016deeper}
I.~Laina, C.~Rupprecht, V.~Belagiannis, F.~Tombari, and N.~Navab.
\newblock Deeper depth prediction with fully convolutional residual networks.
\newblock {\em arXiv preprint arXiv:1606.00373}, 2016.

\bibitem{li2017monocular}
B.~Li, Y.~Dai, and M.~He.
\newblock Monocular depth estimation with hierarchical fusion of dilated cnns
  and soft-weighted-sum inference.
\newblock {\em arXiv preprint arXiv:1708.02287}, 2017.

\bibitem{lin2016refinenet}
G.~Lin, A.~Milan, C.~Shen, and I.~Reid.
\newblock Refinenet: Multi-path refinement networks with identity mappings for
  high-resolution semantic segmentation.
\newblock {\em arXiv preprint arXiv:1611.06612}, 2016.

\bibitem{lin2016efficient}
G.~Lin, C.~Shen, A.~van~den Hengel, and I.~Reid.
\newblock Efficient piecewise training of deep structured models for semantic
  segmentation.
\newblock In {\em CVPR}, 2016.

\bibitem{liu2015deep}
F.~Liu, C.~Shen, and G.~Lin.
\newblock Deep convolutional neural fields for depth estimation from a single
  image.
\newblock In {\em CVPR}, 2015.

\bibitem{liu2014discrete}
M.~Liu, M.~Salzmann, and X.~He.
\newblock Discrete-continuous depth estimation from a single image.
\newblock In {\em CVPR}, 2014.

\bibitem{liu2015semantic}
Z.~Liu, X.~Li, P.~Luo, C.-C. Loy, and X.~Tang.
\newblock Semantic image segmentation via deep parsing network.
\newblock In {\em ICCV}, 2015.

\bibitem{long2015fully}
J.~Long, E.~Shelhamer, and T.~Darrell.
\newblock Fully convolutional networks for semantic segmentation.
\newblock In {\em CVPR}, 2015.

\bibitem{long2015learning}
M.~Long and J.~Wang.
\newblock Learning multiple tasks with deep relationship networks.
\newblock {\em arXiv preprint arXiv:1506.02117}, 2015.

\bibitem{luong2015effective}
M.-T. Luong, H.~Pham, and C.~D. Manning.
\newblock Effective approaches to attention-based neural machine translation.
\newblock {\em arXiv preprint arXiv:1508.04025}, 2015.

\bibitem{misra2016cross}
I.~Misra, A.~Shrivastava, A.~Gupta, and M.~Hebert.
\newblock Cross-stitch networks for multi-task learning.
\newblock In {\em CVPR}, 2016.

\bibitem{mnih2014recurrent}
V.~Mnih, N.~Heess, A.~Graves, et~al.
\newblock Recurrent models of visual attention.
\newblock In {\em NIPS}, 2014.

\bibitem{mousavian2016joint}
A.~Mousavian, H.~Pirsiavash, and J.~Ko{\v{s}}eck{\'a}.
\newblock Joint semantic segmentation and depth estimation with deep
  convolutional networks.
\newblock In {\em 3DV}, 2016.

\bibitem{noh2015learning}
H.~Noh, S.~Hong, and B.~Han.
\newblock Learning deconvolution network for semantic segmentation.
\newblock In {\em ICCV}, 2015.

\bibitem{ranjan2016hyperface}
R.~Ranjan, V.~M. Patel, and R.~Chellappa.
\newblock Hyperface: A deep multi-task learning framework for face detection,
  landmark localization, pose estimation, and gender recognition.
\newblock {\em arXiv preprint arXiv:1603.01249}, 2016.

\bibitem{roymonocular}
A.~Roy and S.~Todorovic.
\newblock Monocular depth estimation using neural regression forest.
\newblock In {\em CVPR}, 2016.

\bibitem{ruder2017sluice}
S.~Ruder, J.~Bingel, I.~Augenstein, and A.~S{\o}gaard.
\newblock Sluice networks: Learning what to share between loosely related
  tasks.
\newblock {\em arXiv preprint arXiv:1705.08142}, 2017.

\bibitem{saxena20083}
A.~Saxena, S.~H. Chung, and A.~Y. Ng.
\newblock 3-d depth reconstruction from a single still image.
\newblock {\em IJCV}, 76(1):53--69, 2008.

\bibitem{saxena2009make3d}
A.~Saxena, M.~Sun, and A.~Y. Ng.
\newblock Make3d: Learning 3d scene structure from a single still image.
\newblock {\em TPAMI}, 31(5):824--840, 2009.

\bibitem{sermanet2013overfeat}
P.~Sermanet, D.~Eigen, X.~Zhang, M.~Mathieu, R.~Fergus, and Y.~LeCun.
\newblock Overfeat: Integrated recognition, localization and detection using
  convolutional networks.
\newblock {\em arXiv preprint arXiv:1312.6229}, 2013.

\bibitem{silberman2012indoor}
N.~Silberman, D.~Hoiem, P.~Kohli, and R.~Fergus.
\newblock Indoor segmentation and support inference from rgbd images.
\newblock In {\em ECCV}, 2012.

\bibitem{simonyan2014very}
K.~Simonyan and A.~Zisserman.
\newblock Very deep convolutional networks for large-scale image recognition.
\newblock {\em arXiv preprint arXiv:1409.1556}, 2014.

\bibitem{teichmann2016multinet}
M.~Teichmann, M.~Weber, M.~Zoellner, R.~Cipolla, and R.~Urtasun.
\newblock Multinet: Real-time joint semantic reasoning for autonomous driving.
\newblock {\em arXiv preprint arXiv:1612.07695}, 2016.

\bibitem{wang2015towards}
P.~Wang, X.~Shen, Z.~Lin, S.~Cohen, B.~Price, and A.~Yuille.
\newblock Towards unified depth and semantic prediction from a single image.
\newblock In {\em CVPR}, 2015.

\bibitem{xia2016zoom}
F.~Xia, P.~Wang, L.-C. Chen, and A.~L. Yuille.
\newblock Zoom better to see clearer: Human and object parsing with
  hierarchical auto-zoom net.
\newblock In {\em ECCV}, 2016.

\bibitem{xu2017learning}
D.~Xu, W.~Ouyang, X.~Alameda-Pineda, E.~Ricci, X.~Wang, and N.~Sebe.
\newblock Learning deep structured multi-scale features using attention-gated
  crfs for contour prediction.
\newblock In {\em NIPS}, 2017.

\bibitem{xu2017multi}
D.~Xu, E.~Ricci, W.~Ouyang, X.~Wang, and N.~Sebe.
\newblock Multi-scale continuous crfs as sequential deep networks for monocular
  depth estimation.
\newblock In {\em CVPR}, 2017.

\bibitem{xu2018structured}
D.~Xu, W.~Wang, H.~Tang, H.~Liu, N.~Sebe, and E.~Ricci.
\newblock Structured attention guided convolutional neural fields for monocular
  depth estimation.
\newblock In {\em CVPR}, 2018.

\bibitem{xu2015show}
K.~Xu, J.~Ba, R.~Kiros, K.~Cho, A.~Courville, R.~Salakhudinov, R.~Zemel, and
  Y.~Bengio.
\newblock Show, attend and tell: Neural image caption generation with visual
  attention.
\newblock In {\em ICML}, 2015.

\bibitem{yao2012describing}
J.~Yao, S.~Fidler, and R.~Urtasun.
\newblock Describing the scene as a whole: Joint object detection, scene
  classification and semantic segmentation.
\newblock In {\em CVPR}, 2012.

\bibitem{yu2015multi}
F.~Yu and V.~Koltun.
\newblock Multi-scale context aggregation by dilated convolutions.
\newblock {\em arXiv preprint arXiv:1511.07122}, 2015.

\bibitem{zhao2016pyramid}
H.~Zhao, J.~Shi, X.~Qi, X.~Wang, and J.~Jia.
\newblock Pyramid scene parsing network.
\newblock {\em arXiv preprint arXiv:1612.01105}, 2016.

\bibitem{zheng2015conditional}
S.~Zheng, S.~Jayasumana, B.~Romera-Paredes, V.~Vineet, Z.~Su, D.~Du, C.~Huang,
  and P.~H. Torr.
\newblock Conditional random fields as recurrent neural networks.
\newblock In {\em ICCV}, 2015.

\bibitem{zhuo2015indoor}
W.~Zhuo, M.~Salzmann, X.~He, and M.~Liu.
\newblock Indoor scene structure analysis for single image depth estimation.
\newblock In {\em CVPR}, 2015.

\end{thebibliography}
}

\end{document}